\documentclass[10pt,twocolumn,letterpaper]{article}

\usepackage{wacv}              

\usepackage{enumitem}
\usepackage{tabularx}   
\usepackage{multirow}   
\usepackage{graphicx}   
\usepackage{booktabs}   
\usepackage{siunitx} 
\usepackage{threeparttable}
\usepackage[accsupp]{axessibility}
\definecolor{wacvblue}{rgb}{0.21,0.49,0.74}
\usepackage[pagebackref,breaklinks,colorlinks,allcolors=wacvblue]{hyperref}

\newcommand\blfootnote[1]{%
  \begingroup
  \renewcommand\thefootnote{}\footnote{#1}%
  \addtocounter{footnote}{-1}%
  \endgroup
}

\def\wacvPaperID{2063} 
\def\confName{WACV}
\def\confYear{2026}

\title{From Lightweight CNNs to SpikeNets: Benchmarking Accuracy–Energy Tradeoffs with Pruned Spiking SqueezeNet}

\author{
Radib Bin Kabir$^{1*}$,
Tawsif Tashwar Dipto$^{1*}$,
Mehedi Ahamed$^{2*}$,
Sabbir Ahmed$^{1}$,
Md Hasanul Kabir$^{1}$ \\
$^{1}$Islamic University of Technology, Dhaka, Bangladesh \\
$^{2}$Southeast University, Dhaka, Bangladesh \\
\tt{
\{radib, tawsiftashwar, sabbirahmed, hasanul\}@iut-dhaka.edu
}\\
\tt{
mehedi.ahamed@seu.edu.bd
}
}

\begin{document}
\maketitle

\begin{abstract}
Spiking Neural Networks (SNNs) are increasingly studied as energy-efficient alternatives to Convolutional Neural Networks (CNNs), particularly for edge intelligence. However, prior work has largely emphasized large-scale models, leaving the design and evaluation of lightweight CNN-to-SNN pipelines underexplored. This work presents the first systematic benchmark of lightweight SNNs obtained by converting compact CNN architectures into spiking networks, where activations are modeled with Leaky-Integrate-and-Fire (LIF) neurons and trained using surrogate gradient descent under a unified setup. 
We construct spiking variants of ShuffleNet, SqueezeNet, MnasNet, and MixNet, along with evaluating them on CIFAR-10, CIFAR-100, and TinyImageNet datasets- measuring accuracy, F1-score, parameter count, computational complexity, and energy consumption. 
Our results show that SNNs can achieve up to \textbf{15.7$\times$} higher energy efficiency than their CNN counterparts while retaining competitive accuracy. Among these, the SNN variant of SqueezeNet consistently outperforms other lightweight SNNs. To further optimize this model, we apply a structured pruning strategy that removes entire redundant fire modules, yielding a pruned architecture, `SNN SqueezeNet-P'. This pruned model improves CIFAR-10 accuracy by \textbf{6\%} and reduces parameters by \textbf{19\%} compared to the original SNN SqueezeNet. Crucially, it narrows the gap with CNN SqueezeNet, achieving nearly the same accuracy (only \textbf{1\%} lower) but with an \textbf{88.1\% reduction} in energy consumption due to sparse spike-driven computations. Together, these findings establish lightweight SNNs as practical, low-power alternatives for edge deployment, highlighting a viable path toward deploying high-performance, low-power intelligence on the edge. Code available at GitHub: \href{https://github.com/MehediAhamed/Pruned-Spiking-SqueezeNet.git}{Pruned-Spiking-SqueezeNet} 
\blfootnote{\hspace{-15pt}$^*$ These authors contributed equally.}
\end{abstract}

\section{Introduction}
\label{sec:intro}
Deep convolutional neural networks (CNNs) \cite{deeplr, meradip} have revolutionized visual recognition; however, their deployment on edge devices and embedded platforms is constrained by high energy consumption and computational demand \cite{Qi2024EnergyCNN}. To address this, lightweight CNNs, such as ShuffleNet\cite{zhang2018shufflenet}, MixNet\cite{tan2019mixnet}, MnasNet\cite{tan2019mnasnet}, and SqueezeNet\cite{iandola2016squeezenet} offer competitive accuracy with significantly fewer parameters and MAC operations, making them more suitable for resource-constrained settings~\cite{zhang2018shufflenet,iandola2016squeezenet,tan2019mixnet}. 

\begin{figure}[t]
    \centering
    \includegraphics[width=0.99\linewidth]{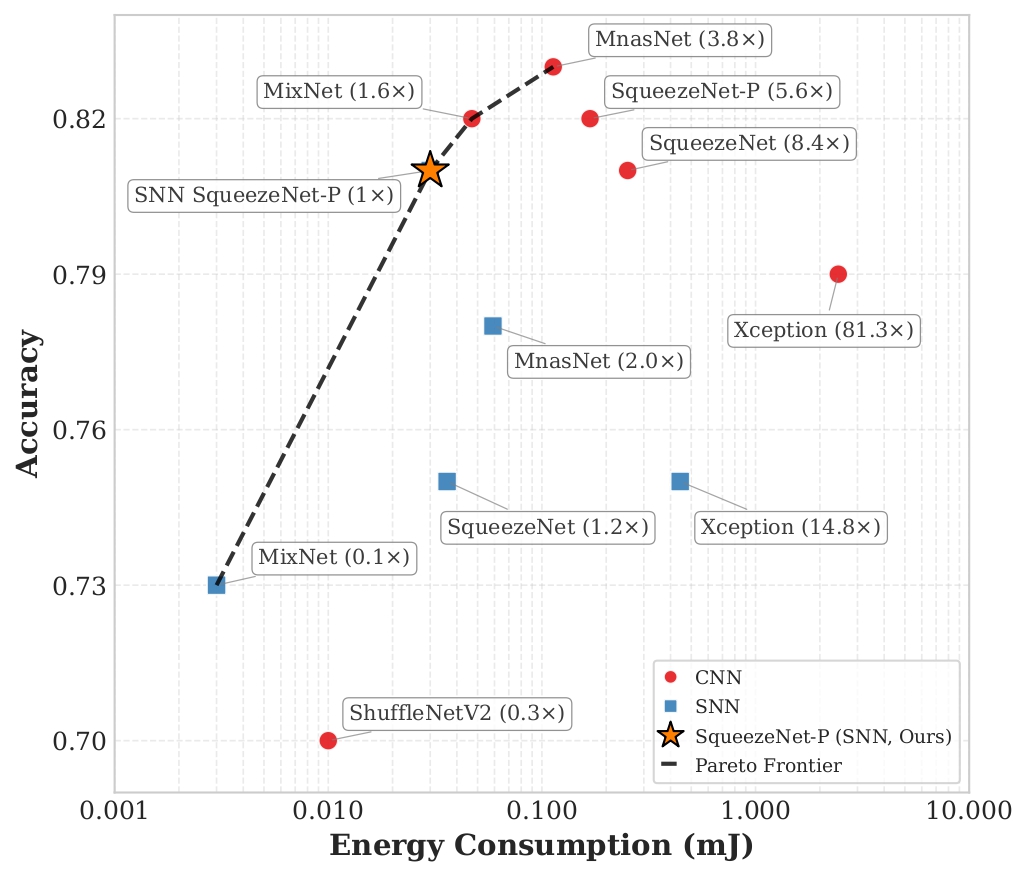}
    \caption{Accuracy-energy trade-off on CIFAR-10. While CNNs achieve strong accuracy, they incur much higher energy costs. SNNs reduce energy but often sacrifice accuracy. Our pruned SNN SqueezeNet-P (orange star) lies on the Pareto frontier, delivering the best balance between accuracy and efficiency.}
    \label{fig:teaser}
\end{figure}

Spiking Neural Networks (SNNs) have emerged as an energy-efficient alternative, leveraging sparse, event-driven communication to reduce energy consumption during inference~\cite{BuildEnergyEfficientSNN2022,SpikingSNNEdge2025}. Neuromorphic hardware, exemplified by chips like Innatera's Pulsar, further illustrates the real-world potential of SNNs to enable always-on sensing with ultra-low power and latency in devices such as smart wearables and IoT sensors~\cite{news20}.

SNNs still struggle to achieve parity in accuracy with their CNN counterparts, especially under tight resource constraints. Conventional CNNs typically occupy the high-accuracy, high-energy region of the design space, while most SNNs fall into the low-energy, low-accuracy quadrant (\figureautorefname~\ref{fig:teaser}). Conversion-based methodologies that transplant pretrained CNN weights into SNNs provide a pragmatic pathway to leverage modern training techniques while gaining spiking efficiency~\cite{ReviewApplicationsSNNs}, yet the literature lacks a systematic evaluation of compact CNN-to-SNN conversions.

In this work, we systematically study spiking versions of lightweight CNNs, including ShuffleNet, MixNet, MnasNet, and SqueezeNet under consistent training as well as evaluation conditions. Central to our contribution is a tailored pruning strategy applied to the Spiking SqueezeNet architecture, designed to eliminate redundant pathways and further optimize efficiency. Through extensive experiments on CIFAR-10, CIFAR-100, and Tiny ImageNet, we evaluate accuracy, F1-score, compute (ACs/MACs), parameter count, and energy consumption estimates grounded in realistic hardware models. Our findings show that carefully optimized spiking networks can capture most of the performance of lightweight CNNs while significantly reducing computational load and energy requirements. As demonstrated by the Pareto-optimal position of our SNN SqueezeNet-P in \figureautorefname~\ref{fig:teaser}, our approach achieves a strong balance between accuracy and energy efficiency, providing practical design guidelines for deploying compact and high-performing SNNs in power-constrained environments.

\section{Related Works}
\label{sec:related_works}

This section provides a comprehensive review of the literature relevant to our study on compact Spiking Neural Networks (SNNs) converted from lightweight Convolutional Neural Networks (CNNs). We organize the discussion into key research areas: ANN-to-SNN conversion methods, lightweight CNN architectures, pruning techniques in neural networks, and performance evaluation of SNNs on standard benchmarks.

\subsection{ANN-to-SNN Conversion Methods}
\label{subsec:conversion_methods}

Converting pre-trained Artificial Neural Networks (ANNs) to Spiking Neural Networks (SNNs) provides a way to leverage established CNN architectures while benefiting from spike-based energy efficiency. Early work by Diehl \textit{et al.} \cite{diehl2015unsupervised} replaced ReLU activations with integrate-and-fire neurons, using rate coding and threshold calibration to maintain activation distributions, achieving competitive MNIST accuracy with reduced energy.

\begin{figure}[t]
    \centering
    \includegraphics[width=.95\linewidth]{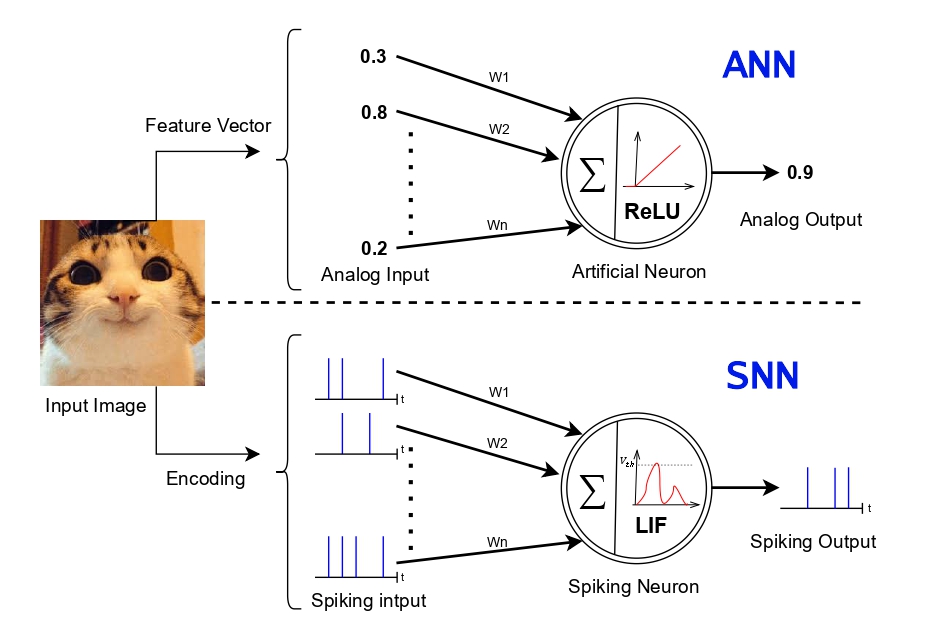}
    \caption{Comparison of neuron-level processing in ANN and SNN (reproduced from \cite{Acharya2020}).}
    \label{fig:ann_vs_snn}
\end{figure}

Subsequent methods addressed deep network conversion challenges. Rueckauer \textit{et al.} \cite{rueckauer2017conversion} proposed normalization techniques for batch-normalized CNNs, achieving 99.1\% on MNIST and 91.6\% on CIFAR-10, albeit requiring up to 2500 time steps. Sengupta \textit{et al.} \cite{sengupta2019going} improved efficiency using signed neurons, enabling VGG networks to reach 91.55\% accuracy on CIFAR-10 with 250 time steps.

Recent advances aim to reduce inference latency while maintaining accuracy. Han \textit{et al.} \cite{han2020rmp} introduced residual membrane potential to accelerate convergence, achieving 93.63\% on CIFAR-10 with 32 time steps. Ngu and Lee \cite{Ngu2022CNNtoSNN} proposed layerwise and channelwise threshold balancing, achieving low-conversion-loss SNNs on MNIST, Fashion-MNIST, and CIFAR-10. \figureautorefname~\ref{fig:ann_vs_snn} illustrates the core difference between ANN and SNN neuron processing: ANN directly processes pixel values, whereas SNN requires spike-based encoding \cite{Acharya2020}.

\subsection{Lightweight CNN Architectures}
\label{subsec:lightweight_cnns}

The development of lightweight CNN architectures has been driven by the need for efficient neural networks suitable for mobile and edge devices. Our study builds upon several influential lightweight architectures that have demonstrated strong performance-efficiency trade-offs.

Howard \textit{et al.} \cite{howard2017mobilenets} introduced MobileNets, which employ depthwise separable convolutions to reduce computational cost while maintaining accuracy. Their approach achieved comparable performance to standard CNNs with 8-9$\times$ fewer parameters and computation. Zhang \textit{et al.} \cite{zhang2018shufflenet} proposed ShuffleNet, incorporating channel shuffle operations and group convolutions to enable efficient information flow across feature channels, achieving superior accuracy-efficiency trade-offs compared to MobileNets on ImageNet.

SqueezeNet, introduced by Iandola \textit{et al.} \cite{iandola2016squeezenet}, employs fire modules consisting of squeeze and expand layers to achieve AlexNet-level accuracy with 50× fewer parameters. The architecture's compact design makes it particularly suitable for deployment on resource-constrained devices.  Tan \textit{et al.} \cite{tan2019mixnet} developed MixNet, which uses mixed depthwise convolutions with different kernel sizes to capture features at multiple scales efficiently, achieving state-of-the-art accuracy on ImageNet among mobile models.

EfficientNet, by Tan and Le \cite{tan2019efficientnet}, introduced compound scaling to uniformly scale network depth, width, and resolution, achieving superior performance across various model sizes. MnasNet \cite{tan2019mnasnet} utilized neural architecture search (NAS) to automatically design mobile-optimized networks.

\subsection{Pruning Techniques in Neural Networks}
\label{subsec:pruning_techniques}

Network pruning has been extensively studied as a method to reduce model complexity while maintaining performance. Our pruning approach for SNNs builds upon established techniques in the CNN domain while considering the unique characteristics of spike-based computation.

Magnitude-based pruning, first introduced by LeCun \textit{et al.} \cite{lecun1990optimal}, removes connections with small weights under the assumption that they contribute minimally to network performance. Han \textit{et al.} \cite{han2015deep} extended this approach to deep networks, achieving $9\times$ compression on AlexNet and $3\times$ on VGG-16 without accuracy loss. Structured pruning methods, such as those proposed by Li \textit{et al.} \cite{li2016pruning}, remove entire filters or channels, enabling hardware acceleration and reducing inference time.

More sophisticated pruning strategies have been developed to consider the importance of network components. Molchanov \textit{et al.} \cite{molchanov2016pruning} proposed variational dropout for automatic relevance determination, while Yang \textit{et al.} \cite{yang2017designing} introduced energy-aware pruning specifically for mobile devices. Liu \textit{et al.} \cite{liu2017learning} demonstrated that pruning during training can be more effective than post-training pruning, leading to better accuracy-efficiency trade-offs.

Shen \textit{et al.} \cite{Shenetal} introduced ESL-SNNs, an evolutionary sparse learning method that prunes and regrows connections during training. Shi \textit{et al.} \cite{shi2024towards} jointly pruned weights and neurons under an energy model to significantly reduce synaptic operations (SOPs). Later, Shen \textit{et al.} \cite{Shenetal2} proposed a two-stage sparse learning scheme using the PQ index to set pruning ratios. Roy \textit{et al.} \cite{roy2019towards} showed that SNNs inherently produce sparse activations, enabling aggressive pruning with minimal accuracy loss.

\subsection{Evaluation of SNNs on Standard Benchmarks}
\label{subsec:snn_benchmarks}

Spiking Neural Networks (SNNs) have attracted increasing interest as energy-efficient alternatives to conventional deep neural networks, owing to their event-driven computation and sparse activations \cite{Neftci2019SurrogateGradient, Acharya2020, roy2019towards}. Early models relied on unsupervised STDP rules for digit recognition \cite{diehl2015unsupervised, kheradpisheh2018stdp}, while later works introduced surrogate gradient learning, enabling deeper and more accurate architectures \cite{sengupta2019going, han2020rmp, fang2021incorporating}. Conversion-based methods map trained CNNs into SNNs and achieve strong results on benchmarks \cite{rueckauer2017conversion, Ngu2022CNNtoSNN}, but typically require high firing rates and longer inference windows. Direct training approaches, by contrast, aim to bridge the performance gap under strict latency constraints \cite{zheng2021going, deng2022temporal}.

In parallel, lightweight CNNs such as SqueezeNet \cite{iandola2016squeezenet}, ShuffleNet \cite{zhang2018shufflenet}, and MixNet \cite{tan2019mixnet} demonstrate how structural innovations depthwise separable convolutions, channel shuffle, or mixed kernels, can drastically reduce computational cost. Coupled with pruning and compression techniques \cite{han2015deep, li2016pruning, yang2017designing}, these models offer compact yet accurate alternatives for resource-limited environments.

On standard benchmarks, computationally heavy CNNs such as ResNet and VGG routinely achieve $>$95\% on CIFAR-10 and $>$75\% on CIFAR-100, setting the bar for accuracy. In contrast, directly trained SNNs typically report $\sim$85–90\% on CIFAR-10 and $\sim$65–70\% on CIFAR-100 \cite{zheng2021going, deng2022temporal}, reflecting the inherent trade-off between accuracy and efficiency. Recent studies have begun combining efficiency-driven CNN principles with SNN training \cite{BuildEnergyEfficientSNN2022, SpikingSNNEdge2025}, highlighting the promise of lightweight spiking architectures for neuromorphic deployment despite the remaining performance gap.

\subsection{Gaps and Contributions}
\label{subsec:gaps_contributions}

Despite significant progress in ANN-to-SNN conversion, several important gaps remain. Prior work has predominantly focused on large-scale networks such as VGG and ResNet, while lightweight architectures have received limited attention. Furthermore, the effects of architectural pruning on converted SNNs have not been systematically studied across diverse datasets. In this work, we address these gaps with three main contributions:

\begin{enumerate}[label=(\roman*)]
    \item We present the first comprehensive benchmark of several state-of-the-art lightweight CNNs, including ShuffleNet, MnasNet, MixNet, and SqueezeNet, converted into their SNN counterparts, evaluated across multiple standard datasets.
    \item Based on our benchmarking, we identify SNN SqueezeNet as the best-performing model among the evaluated architectures. We then introduce a pruning strategy tailored for SNN SqueezeNet, improving its accuracy while maintaining energy efficiency and reducing the gap with the original CNN. 
    \item We provide a detailed comparison of CNNs and their SNN equivalents in terms of both accuracy and energy consumption to understand the accuracy-energy trade-off, offering practical insights for efficient deployment on resource-constrained hardware.
\end{enumerate}

Collectively, these contributions advance the understanding and practical deployment of lightweight SNNs.

\section{Methodology}
\label{sec:method}
In this study, we present a methodology for developing and evaluating lightweight SNNs for image classification tasks. Our approach focuses on systematically converting conventional lightweight CNNs, specifically ShuffleNet, SqueezeNet, MnasNet, and MixNet, into SNNs and training them under controlled conditions. Additionally, for the SNN SqueezeNet, which consistently demonstrated the highest accuracy among the converted models, we further explore a pruning strategy to improve both parameter efficiency and energy consumption.

\subsection{Datasets}
We evaluated our SNN models on three benchmark datasets commonly used in image classification research to ensure robustness across varying complexities and class distributions:

\begin{itemize}
\item \textbf{CIFAR-10 \cite{cifar_krizhevsky}}: This dataset consists of 60,000 color images of $32\times32$ size, divided into 10 classes (e.g., airplanes, automobiles, birds), with 50,000 images for training and 10,000 for testing. Each class is balanced with 6,000 images.

\item \textbf{CIFAR-100 \cite{cifar_krizhevsky}}: Similar to CIFAR-10 in image size and total count (60,000 images), but with 100 fine-grained classes grouped into 20 superclasses. It presents a more challenging task due to increased intra-class variability and inter-class similarity.

\item \textbf{TinyImageNet \cite{tinyimagenet}}: A subset of the larger ImageNet dataset, comprising 200 classes with 100,000 $64\times64$ color images for training (500 per class), 10,000 for validation, and 10,000 for testing. Images were resized to $32\times32$ to match the input dimensions of our models, and standard normalization (mean=[0.485, 0.456, 0.406], std=[0.229, 0.224, 0.225]) was applied.
\end{itemize}

The datasets were chosen for their progressive difficulty, allowing to assess the scalability of our SNN approach from simple to more complex classification scenarios.

\subsection{SNN Conversion}

\begin{figure*}[t]
\centering
\includegraphics[width=\textwidth,height=0.33\textheight,keepaspectratio]{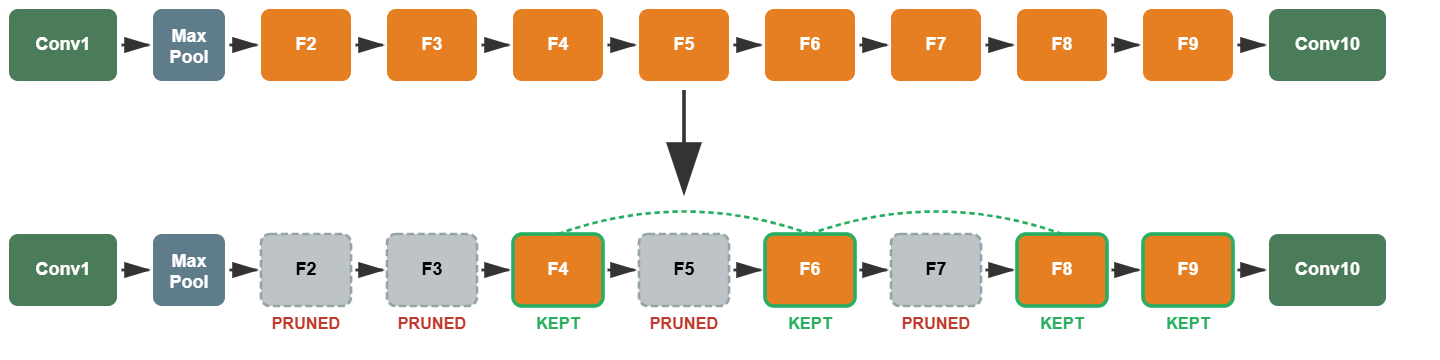}
\caption{
Pruning strategy for SNN SqueezeNet. Each `Fi' denotes a Fire (firing) module `i'. The top row shows the original SNN SqueezeNet with all Fire modules active. The bottom row illustrates the pruned SNN SqueezeNet, where ineffective Fire modules (F2, F3, F5, F7) are removed, and only selected modules (F4, F6, F8, F9) are kept. The dotted green line indicates that the retained Fire modules maintain their functional connections, ensuring information flow through the pruned network.
}
\label{fig:methodology}
\end{figure*}

To bridge the gap between ANNs and SNNs, we adopted a conversion-based approach starting from a pre-trained ANN model. Specifically:

\begin{itemize}

\item \textbf{Neuron Replacement}: ReLU activations in the ANN were replaced with Leaky Integrate-and-Fire (LIF) neurons, a biologically plausible spiking model. The LIF neuron dynamics are governed by the following differential equation:
$$\tau \frac{dV}{dt} = -(V - V_{\text{rest}}) + I(t) \cdot R,$$

where $V$ is the membrane potential, $\tau$ is the time constant (set to 2.0), $V_{\text{rest}}$ is the resting potential (0.0), $I(t)$ is the input current and $R$ is the resistance (1.0). A spike is emitted when $V$ exceeds a threshold $V_{\text{th}}$ (1.0), followed by a reset to $V_{\text{reset}}$ (0.0). Leakage was incorporated with a factor of 0.1 to mimic realistic neuronal behavior and stabilize training.

\item \textbf{Input Encoding:} We fed the pixel intensities directly into the SNN as constant current inputs. This method does not rely on generating spikes over multiple time steps, so the energy cost per image is based on a single forward processing of the network. We avoided Poisson encoding to keep the input deterministic and efficient.

\end{itemize}
This conversion ensures compatibility with backpropagation while preserving the event-driven nature of SNNs, reducing energy consumption compared to traditional ANNs.

\subsection{Model Architecture}
To identify an optimal base architecture for our SNN, we first created SNN counterparts for several popular lightweight CNN models: ShuffleNet, XceptionNet, MNASNet, MixNet, and SqueezeNet. These were converted using the aforementioned SNN conversion process and evaluated on the CIFAR-10 dataset for initial performance comparison. Empirical results showed that the SNN variant of SqueezeNet outperformed the others in terms of accuracy, spike efficiency, and parameter count, making it the selected base model for further refinement.
Our base architecture is thus a spiking variant of SqueezeNet, chosen for its lightweight design and efficiency in resource-constrained environments. SqueezeNet employs fire modules that consist of a squeeze layer ($1\times1$ convolutions) followed by expand layers (mix of $1\times1$ and $3\times3$ convolutions), achieving AlexNet-level accuracy with significantly fewer parameters.

\emph{\textbf{SNN SqueezeNet}}: We adapted SqueezeNet by integrating LIF neurons into each convolutional and fully connected layer. The network structure includes an initial convolution (conv1), followed by max-pooling, eight fire modules (fire2 to fire9), another max-pooling, and a final global average pooling leading to a softmax classifier. Spiking occurs across $T$ time steps, with temporal accumulation of membrane potentials.

\emph{\textbf{Pruning Technique:}} 
To optimize Spiking SqueezeNet, we employed a structured pruning technique targeting specific \texttt{fire} and \texttt{max-pooling} layers. Guided by ablation studies (\tableautorefname~\ref{tab:ablation_pruning}), we systematically removed layers contributing minimally to performance. This led to a leaner architecture with improved energy efficiency and reduced parameter count, while preserving classification accuracy. The final configuration shown in \figureautorefname~\ref{fig:methodology} retains \texttt{fire4}, \texttt{fire6}, \texttt{fire8}, and \texttt{fire9}, balancing accuracy and computational economy for energy-constrained applications.

\subsection{Experimental Setup and Training Strategy}

All models, including both CNN and SNN variants, were trained from scratch for 120 epochs using the Adam optimizer \cite{kingma2015adam} with an initial learning rate of 0.001, decayed by a factor of 0.1 at epochs 50 and 100. A batch size of 12 was used due to memory constraints. Early stopping with a patience of 10 epochs was applied based on validation accuracy. To ensure a fair comparison of computational cost and energy efficiency, no pretrained weights were used for either SNN or CNN models, avoiding discrepancies in total operation count and spike activity. Spike rates were monitored for SNNs, targeting an average firing rate below 0.3 per neuron per time step. All experiments were conducted on NVIDIA Tesla P100 GPUs using PyTorch and the SpikingJelly framework.

Training SNNs presents challenges due to the non-differentiable nature of spiking functions. We addressed this using the following techniques:

\begin{itemize}
    \item \textbf{Surrogate Gradient Descent:} To enable backpropagation through the Heaviside step function used in spiking neurons, we employ a differentiable surrogate gradient during the backward pass \cite{Neftci2019SurrogateGradient}. Specifically, we use the arctangent-based surrogate function implemented in SpikingJelly's \texttt{surrogate.ATan()}~\cite{Fang2023SpikingJelly} :
    \begin{equation}
    \label{eq:sigma_prime}
    \sigma'(u) = \frac{\alpha}{2} \cdot \frac{1}{1 + \left( \frac{\pi \alpha u}{2} \right)^2},
    \end{equation}
    where \( u = v - V_{th} \) is the normalized membrane potential, \( v \) is the membrane potential, \( V_{th} \) is the firing threshold and \( \alpha \) is a scaling hyperparameter controlling the gradient's steepness. This surrogate gradient approximates the derivative of the Heaviside function, enabling end-to-end training via backpropagation through time (BPTT) while preserving discrete spike generation (\( s = H(v - V_{th}) \)) in the forward pass.
    
    \item \textbf{Loss Functions:} The primary objective was multi-class classification, optimized using Cross-Entropy Loss:
    
    \begin{equation}
    \label{eq:cross_entropy}
    \mathcal{L}_{\text{CE}} = -\sum_{i=1}^{C} y_i \log(\hat{y}_i),
    \end{equation}
    where \(y_i\) is the ground truth and \(\hat{y}_i\) is the predicted probability averaged over time steps. To mitigate vanishing gradients in deep SNNs, we introduce a Gradient-Aware Loss that penalizes low gradient magnitudes in intermediate layers:
    \begin{equation}
    \label{eq:gradaware_loss}
    \mathcal{L}_{\text{GA}} = \lambda \sum_{l} \left( 1 - \frac{||\nabla_{W_l} \mathcal{L}_{\text{CE}}||_2}{||\nabla_{W_l} \mathcal{L}_{\text{CE}}||_2 + \epsilon} \right),
    \end{equation}
    where \( \nabla_{W_l} \mathcal{L}_{\text{CE}} \) is the gradient of the cross-entropy loss with respect to the weights \( W_l \) of layer \( l \), \( ||\cdot||_2 \) denotes the L2 norm, \( \lambda \) is a weighting factor and \( \epsilon \) ensures numerical stability. The total loss is:
    \begin{equation}
    \label{eq:total_loss}
    \mathcal{L} = \mathcal{L}_{\text{CE}} + \mathcal{L}_{\text{GA}}.
    \end{equation}
\end{itemize}

This methodology provides a controlled and comprehensive framework for comparing CNN and SNN models, balancing biological fidelity, energy efficiency, and practical performance while ensuring fair and reproducible evaluation.

\subsection{Evaluation Metrics}
To comprehensively assess the performance of Spiking Neural Networks (SNNs) and their CNN counterparts, we employ a diverse set of evaluation metrics:

\begin{itemize}
    \item \textbf{Accuracy (Acc)}: The top-1 classification accuracy, which measures the proportion of correctly predicted samples over the total test set.
    
    \item \textbf{F1-Score (F1)}: The harmonic mean of precision and recall, providing a balanced evaluation for datasets with potential class imbalance.
    
    \item \textbf{Parameters (Params)}: The total number of trainable parameters, reported in thousands (K) or millions (M), representing the model's memory footprint.
    
    \item \textbf{Multiply-Accumulate Operations (MACs)}: MAC denotes one multiplication and one accumulation and is used for all convolution and linear operations in CNNs. For CNNs, computational complexity is dominated by MAC operations. We estimate MACs using the \texttt{ptflops}~\cite{ptflops} library.
    
    \item \textbf{Accumulate Operations (ACs)}: AC refers to an addition-only event triggered by a spike in SNNs, where no multiplication is performed because weights are applied only to spike events. Unique to SNNs, ACs represent spike-driven accumulation events, which incur lower energy costs due to their sparse and event-driven nature.
    
    \item \textbf{Energy Consumption}: Energy efficiency is critical for edge deployment. Following Horowitz~\cite{Horowitz2014}, the energy cost is approximated as:
    \begin{equation}
        E_{\text{total}} = (N_{\text{AC}} \times 0.9\ \text{pJ}) + (N_{\text{MAC}} \times 4.6\ \text{pJ}),
    \end{equation}
    where $N_{\text{AC}}$ and $N_{\text{MAC}}$ denote the total number of accumulate and multiply-accumulate operations, respectively, and \texttt{pJ} represents picojoules. For CNNs, only MACs are considered, while for SNNs both ACs and MACs are accounted for using the \texttt{syOps}~\cite{chen2023training} library.
\end{itemize}

This combination of accuracy, F1-score, model complexity, and energy estimates enables a holistic evaluation of trade-offs between predictive performance and computational efficiency.

\section{Result Analysis}
\label{sec:result}

\subsection{Performance Comparison Among SNN Models}

\begin{table*}[t]
\centering
\caption{Performance comparison of SNN architectures across datasets}
\label{tab:performance}
\small
\begin{tabular}{@{}l l c c c c c c@{}}
\toprule
\textbf{Dataset} & \textbf{Network} & \textbf{Acc} & \textbf{F1} & \textbf{AC (k/M)} & \textbf{MAC (k/M)} & \textbf{Params (k/M)} & \textbf{Energy (mJ)} \\
\midrule
\multirow{6}{*}{CIFAR10} 
 & SNN-ShuffleNetV2    & 0.40 & 0.40 & \textbf{93}    & \textbf{894}    & 911.0  & 0.0042 \\
 & SNN-Xception        & 0.75 & 0.75 & 1510  & 96300  & 1025.0 & 0.4444 \\
 & SNN-MnasNet         & 0.78 & 0.77 & 3430  & 12130  & 151.8  & 0.0589 \\
 & SNN-MixNet          & 0.73 & 0.73 & 523   & 506.9  & \textbf{84.90}  & \textbf{0.0028} \\
 & SNN-SqueezeNet      & 0.75 & 0.74 & 16610 & 4470   & 736.5  & 0.0355 \\
 & SNN-SqueezeNet-P (ours) & \textbf{0.81} & \textbf{0.81} & 11200 & 4230   & 599.1  & 0.0295 \\
\midrule
\multirow{6}{*}{CIFAR100} 
 & SNN-ShuffleNetV2    & 0.12 & 0.1 & \textbf{163}   & 894.6  & 1000.0 & 0.0043 \\
 & SNN-Xception        & 0.36 & 0.34 & 1780  & 96310  & 1025.0 & 0.4446 \\
 & SNN-MnasNet         & 0.49 & 0.48 & 3050  & 12930  & 163.5  & 0.0622 \\
 & SNN-MixNet          & 0.44 & 0.43 & 862   & \textbf{506.9}  & \textbf{108.0}  & \textbf{0.0023} \\
 & SNN-SqueezeNet      & 0.47 & 0.47 & 20790 & 4470   & 782.6  & 0.0393 \\
 & SNN-SqueezeNet-P (ours) & \textbf{0.54} & \textbf{0.53} & 13490 & 4230   & 645.3  & 0.0316 \\
\midrule
\multirow{6}{*}{Tiny ImageNet} 
 & SNN-ShuffleNetV2    & 0.11 & 0.09 & 642   & 3780   & 1110.0 & 0.0180 \\
 & SNN-Xception        & 0.23 & 0.21 & 6440  & 385630 & 10460.0& 1.7797 \\
 & SNN-MnasNet         & 0.38 & 0.37 & 12040 & 51690  & 176.4  & 0.2486 \\
 & SNN-MixNet          & 0.34 & 0.33 & \textbf{2000}  & \textbf{203.0}  & \textbf{133.7}  & \textbf{0.0093} \\
 & SNN-SqueezeNet      & 0.41 & 0.40 & 80094 & 17880  & 833.9  & 0.1543 \\
 & SNN-SqueezeNet-P (ours) & \textbf{0.45} & \textbf{0.44} & 51970 & 16920  & 696.6  & 0.1246 \\
\bottomrule
\end{tabular}%
\end{table*}

We evaluated six SNN architectures, \textit{ShuffleNetV2}, \textit{Xception}, \textit{MnasNet}, \textit{MixNet}, \textit{SqueezeNet} and a pruned \textit{SqueezeNet-P} on the CIFAR-10, CIFAR-100, and Tiny ImageNet (200 classes) datasets. All experiments were conducted using a fixed random seed (42). The finding are listed in \tableautorefname~\ref{tab:performance}.

\subsubsection{Performance on CIFAR-10}
On CIFAR-10, the pruned \textit{SNN-SqueezeNet-P} achieved the best overall performance with an accuracy and F1 score of $0.81$, outperforming the baseline SqueezeNet by $6\%$. This gain highlights the effectiveness of pruning, which not only removed redundant connections but also improved generalization by reducing overfitting. Importantly, pruning also reduced energy consumption from $0.0355$ mJ to $0.0295$ mJ, showing that structured sparsity can yield accuracy improvements while lowering computation.

Among the lightweight models, \textit{MixNet} consumed the least energy ($0.0028$ mJ) due to its extremely low parameter count (84.9K), but its accuracy (0.73) lagged behind deeper architectures. \textit{MnasNet} (accuracy 0.78) struck a favorable trade-off, balancing competitive accuracy with moderate energy ($0.0589$ mJ). \textit{Xception}, despite its depth and parameter size, did not surpass smaller models, suggesting that heavily parameterized architectures do not fully translate their potential into SNNs, where spiking dynamics introduce additional constraints. Overall, architectural modifications in SqueezeNet proved to be more beneficial than simply scaling model depth or width.

\subsubsection{Performance on CIFAR-100}
As expected, all models experienced accuracy drops on CIFAR-100 due to the higher inter-class similarity and the 10$\times$ increase in classes. Nevertheless, pruned \textit{SNN-SqueezeNet-P} again achieved the best accuracy (0.54) and F1 score (0.53), outperforming baseline SqueezeNet by $7\%$. This demonstrates that pruning preserved salient features while mitigating overfitting, which becomes more critical in complex datasets.

Interestingly, larger models such as \textit{Xception} (accuracy 0.36) incurred high computational costs ($0.4446$ mJ) without yielding proportional gains, highlighting inefficiency under spiking constraints. In contrast, \textit{MixNet} achieved extremely low energy consumption ($0.0023$ mJ) but at the cost of moderate accuracy (0.44), suggesting its suitability only for ultra-low-power tasks where accuracy can be compromised. \textit{MnasNet} (accuracy 0.49) again demonstrated strong robustness, making it a balanced choice across accuracy and efficiency. These results indicate that scaling up architecture size does not necessarily help SNNs on fine-grained classification tasks, where pruning-based compact designs provide better resilience.

\subsubsection{Performance on Tiny ImageNet}
Tiny ImageNet posed the greatest challenge with 200 classes, leading to significant accuracy reductions across all models. Despite this, pruned \textit{SNN-SqueezeNet-P} maintained its superiority, achieving accuracy of $0.45$ and F1 score of $0.44$, which is a $4\%$ improvement over the baseline SqueezeNet. This consistent advantage across datasets emphasizes that pruning not only reduces energy but also improves robustness to increasing class diversity.

From an efficiency perspective, \textit{MixNet} ($0.0093$ mJ) and \textit{ShuffleNetV2} ($0.0180$ mJ) remained the most energy-efficient, but their accuracies (0.34 and 0.11, respectively) were too low for practical deployment. \textit{Xception}, despite its large capacity, failed to adapt effectively, yielding only 0.23 accuracy at an extremely high energy cost ($1.7797$ mJ), indicating inefficiency for large-scale class recognition in SNN deployments. This highlights the diminishing returns of deep, high-capacity architectures when adapted to the spiking domain for large-scale recognition tasks.


\subsubsection{Overall Trends Across Datasets}
Across all datasets, pruned \textit{SNN-SqueezeNet-P} achieved the best trade-off between accuracy and efficiency, outperforming the baseline by $+6\%$ on CIFAR-10, $+7\%$ on CIFAR-100, and $+4\%$ on Tiny ImageNet while also reducing energy use. Pruning proved effective in removing redundancy and improving generalization, whereas extremely lightweight models like \textit{MixNet} and \textit{ShuffleNetV2} offered very low energy consumption but poor accuracy. In contrast, mid-sized models such as \textit{MnasNet} provided a strong balance, highlighting that compact, pruned designs are more robust and efficient for SNNs, especially as task complexity increases.

\subsection{SNN vs. CNN Comparison}
\label{sec:cnn_snn_tradeoff}

\begin{table}[h]
\centering
\caption{Accuracy and energy comparison between CNN and SNN models on CIFAR-10.}
\label{tab:cnn_snn_tradeoff}
\small
\begin{tabular}{lcc|c|cc|c}
\toprule
\multirow{2}{*}{\textbf{Model}} & \multicolumn{2}{c|}{\textbf{Acc (\%)}} & \multirow{2}{*}{$\Delta \text{A}$} & \multicolumn{2}{c|}{\textbf{Energy (mJ)}} & \multirow{2}{*}{$\mathbf{\eta_E}$} \\
\cmidrule(lr){2-3} \cmidrule(lr){5-6}
 & CNN & SNN & & CNN & SNN & \\
\midrule
ShuffleNetV2   & 70 & 40 & 30 & \textbf{0.010} & 0.004 & 2.5  \\
Xception       & 79 & 75 & 4  & 2.44  & 0.444 & 5.5  \\
MnasNet        & \textbf{83} & 78 & 5  & 0.113 & 0.059 & 1.9  \\
MixNet         & 82 & 73 & 9  & 0.047 & \textbf{0.003} & 15.7 \\
SqueezeNet     & 81 & 75 & 6  & 0.252 & 0.036 & 7.0  \\
SqueezeNet-P   & 82 & \textbf{81} & 1  & 0.168 & 0.030 & 5.6  \\
\bottomrule
\end{tabular}
\begin{tablenotes}
\footnotesize
\item $E$ = energy consumption measured in millijoules (mJ).  
\item $\mathbf{\eta_E} = E_\text{CNN}/E_\text{SNN}$; indicates how many times more energy the CNN model consumes compared to the corresponding SNN model.
\item $\Delta \text{A} = \text{Accuracy}_\text{CNN} - \text{Accuracy}_\text{SNN}$
\end{tablenotes}
\end{table}

Table~\ref{tab:cnn_snn_tradeoff} compares CNN models with their SNN counterparts on CIFAR-10 in terms of accuracy and energy consumption. For this comparison, the standard CNN baselines (ShuffleNetV2, Xception, MnasNet, MixNet, and SqueezeNet) are used in their original unpruned form, while \textit{SqueezeNet-P} is the only model where pruning is applied to both the CNN and SNN versions using the same pruned architecture. This ensures a fair comparison for the pruned variant while keeping the other baselines consistent with prior work. The results show that CNNs generally achieve slightly higher accuracy, but their energy requirements are substantially larger. For example, \textit{MnasNet} records $0.83$ accuracy in the CNN domain versus $0.78$ in the SNN, while its energy reduces by more than half ($0.113$ mJ to $0.059$ mJ). Similarly, \textit{MixNet} and \textit{SqueezeNet} achieve $15.7\times$ and $7\times$ lower energy usage, respectively, when converted to SNNs. An exception is \textit{ShuffleNetV2}, where the SNN version suffers a large accuracy drop (0.70 to 0.40) due to the reliance on channel shuffling and group convolutions, which interact poorly with sparse spiking activity. Most notably, our pruned \textit{SqueezeNet-P} achieves nearly identical accuracy in its CNN and SNN forms ($0.82$ vs.\ $0.81$) while reducing energy consumption by over $5.6\times$, demonstrating that pruning can produce architectures that are simultaneously compact, accurate, and energy-efficient under spiking dynamics. These findings underscore that SNNs provide a compelling trade-off: small accuracy losses are offset by substantial energy savings, making them attractive for low-power neuromorphic systems.

\subsection{Ablation Study: Structured Pruning of SNN-SqueezeNet}

To investigate the relative importance of different Fire modules, we conducted a structured ablation study on CIFAR-10 by pruning entire modules according to systematic schedules. Our analysis proceeded in stages: (1) testing the importance of early (\textit{head}) vs. late (\textit{tail}) modules, (2) evaluating alternating retention patterns, and (3) refining the alternating strategy by fixing consistently important modules and searching for the optimal fourth block. Table~\ref{tab:ablation_pruning} summarizes the results.

\begin{table}[h]
\centering
\caption{Ablation study on pruning Fire modules from SNN-SqueezeNet on CIFAR-10. (\checkmark) indicates the module is retained, (--) indicates it is pruned. Rows are grouped by pruning schedule.}
\label{tab:ablation_pruning}
\small
\resizebox{\linewidth}{!}{
\begin{tabular}{l|cccccccc|c|c|c}
\toprule
\textbf{Schedule} & \textbf{F2} & \textbf{F3} & \textbf{F4} & \textbf{F5} & \textbf{F6} & \textbf{F7} & \textbf{F8} & \textbf{F9} & \textbf{Acc.} & \textbf{Params} & \textbf{E (mJ)} \\
\midrule
\multicolumn{12}{l}{\textbf{Baseline}} \\
Full Model        & \checkmark & \checkmark & \checkmark & \checkmark & \checkmark & \checkmark & \checkmark & \checkmark & 0.75 & 736,500 & 0.0355 \\
\midrule
\multicolumn{12}{l}{\textbf{Head-Heavy Pruning}} \\
Head-1            & \checkmark & \checkmark & \checkmark & \checkmark & -- & -- & -- & \checkmark & 0.77 & 559,080 & 0.0241 \\
Head-2            & \checkmark & \checkmark & \checkmark & \checkmark & -- & -- & \checkmark & -- & 0.77 & 376,220 & 0.0257 \\
\midrule
\multicolumn{12}{l}{\textbf{Tail-Heavy Pruning}} \\
Tail-1            & \checkmark & -- & -- & \checkmark & \checkmark & \checkmark & \checkmark & \checkmark & 0.78 & 673,720 & 0.0304 \\
Tail-2            & \checkmark & -- & \checkmark & -- & \checkmark & \checkmark & \checkmark & \checkmark & 0.79 & 621,660 & 0.0291 \\
\midrule
\multicolumn{12}{l}{\textbf{Alternating Pruning}} \\
Alt-1             & \checkmark & -- & \checkmark & -- & \checkmark & -- & \checkmark & -- & 0.79 & 362,030 & 0.0260 \\
Alt-2             & -- & \checkmark & -- & \checkmark & -- & \checkmark & -- & \checkmark & 0.76 & 363,050 & \textbf{0.0226} \\
\midrule
\multicolumn{12}{l}{\textbf{Refined Alternating}} \\
Ref-1             & -- & -- & \checkmark & -- & \checkmark & -- & \checkmark & \checkmark & \textbf{0.81} & 599,130 & 0.0295 \\
Ref-2             & \checkmark & -- & \checkmark & \checkmark & \checkmark & -- & \checkmark & \checkmark & 0.79 & 735,950 & 0.0279 \\
\bottomrule
\end{tabular}
}
\end{table}

The results follow a clear progression. First, \textbf{Head-Heavy pruning} (removing later layers) retained competitive accuracy (77\%) while delivering the lowest energy consumption (0.0241 mJ), indicating that shallow spiking features are sufficient when efficiency is prioritized. Second, \textbf{Tail-Heavy pruning} preserved later layers and consistently outperformed head-heavy variants (up to 79\%), confirming that deep spiking modules carry stronger discriminative power. Third, in the \textbf{Alternating pruning} experiments, we observed that models with exactly four firing modules performed best. In particular, configurations retaining \{F4, F6, F8\} consistently outperformed others, highlighting these as critical processing blocks. However, the fourth retained module significantly influenced results: when chosen from the head side (Alt-1, F2), accuracy reached 79\%, whereas retaining a tail-side block (Alt-2, F9) offered better efficiency (0.0226 mJ) but lower accuracy (76\%). Building on this insight, we performed \textbf{Refined Alternating pruning}, fixing \{F4, F6, F8\} and varying the fourth block. Since earlier results indicated that tail modules are generally more important than head modules, we paired them with F9. This configuration (Ref-1) achieved the highest accuracy overall (81\%) with reduced energy (0.0295 mJ), demonstrating that the optimal trade-off emerges when late-stage modules are emphasized.  

\begin{figure}[t]
    \centering
    \begin{subfigure}{0.49\linewidth}
        \centering
        \includegraphics[width=\linewidth]{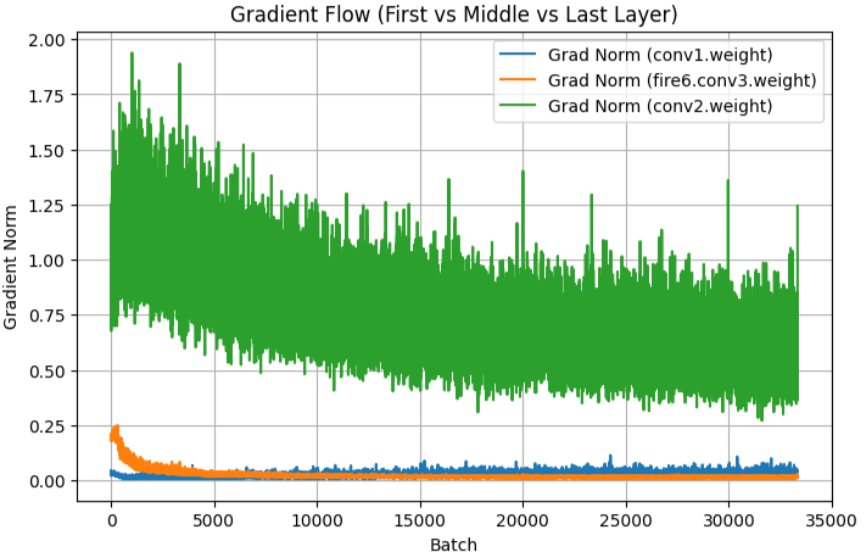}
        \caption{Before pruning}
    \end{subfigure}
    \hfill
    \begin{subfigure}{0.49\linewidth}
        \centering
        \includegraphics[width=\linewidth]{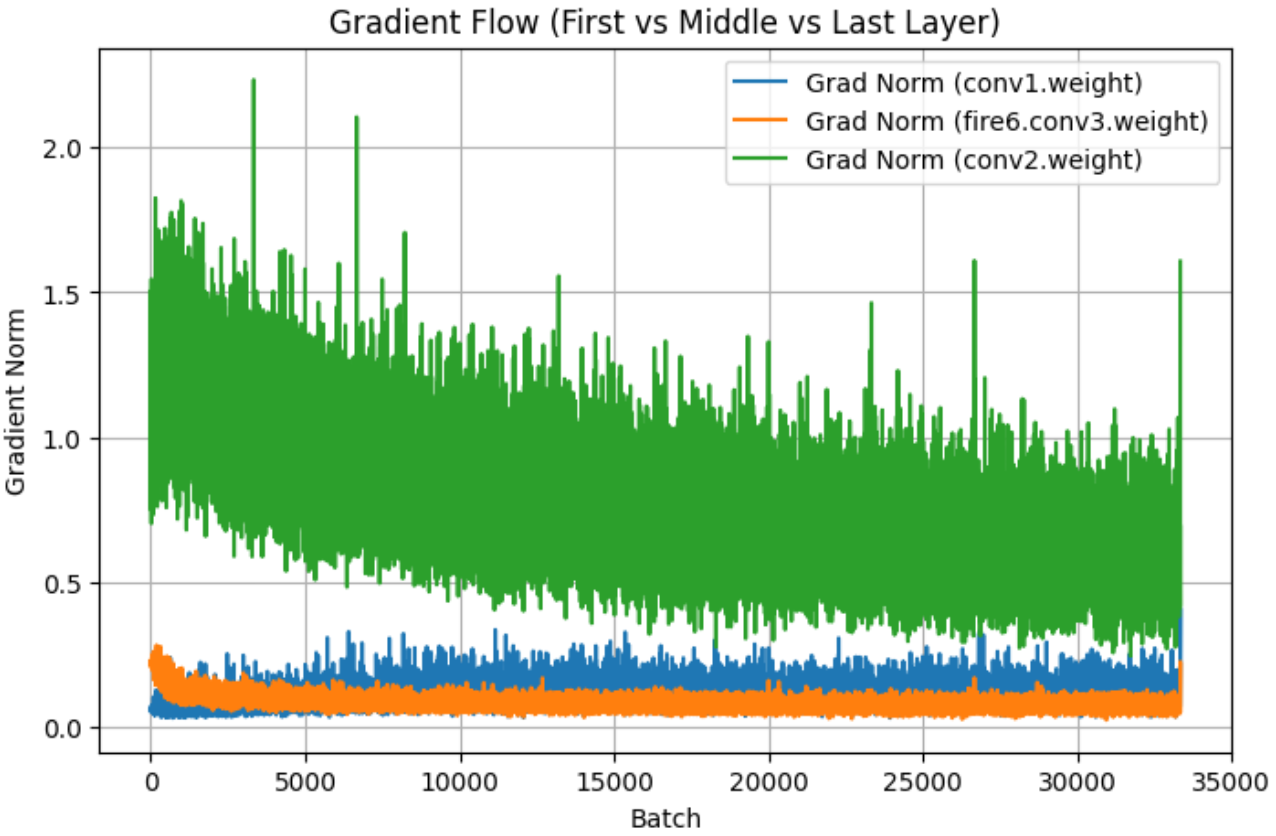}
        \caption{After pruning}
    \end{subfigure}
    \caption{Gradient flow visualization of SNN-SqueezeNet before and after structured pruning. The pruning process removes modules that accumulate vanishing gradients and restores effective gradient propagation.}
    \label{fig:grad_flow}
\end{figure}

This ablation reveals three key principles: (1) late spiking modules are more critical than early ones, (2) models with exactly four firing modules strike the best accuracy–efficiency balance, and (3) the combination \{F4, F6, F8, F9\} forms the most effective subset, providing complementary spike pathways that maximize discriminative power while controlling energy usage. This finding is further reinforced by the gradient-flow behaviour observed during training (Fig.~\ref{fig:grad_flow}): before pruning, the middle layers’ gradients collapse close to zero, indicating severe gradient starvation, whereas pruning restores healthier gradient propagation, strengthening both early and bottleneck layers. This improved balance in gradient flow aligns directly with the ablation results, explaining why keeping deeper modules while selectively pruning others leads to the strongest accuracy–efficiency trade-off.



\section{Conclusion}
\label{sec:conclusion}

In this work, we conducted a systematic evaluation of lightweight CNN-to-SNN conversions across CIFAR-10, CIFAR-100, and Tiny ImageNet datasets, profiling accuracy, complexity, and energy efficiency. Among the tested architectures, SqueezeNet consistently offered the best balance of performance and efficiency, and our proposed pruned variant further improved accuracy by up to $7\%$ while lowering energy consumption. The ablation study revealed that not all modules contribute equally under spiking dynamics, and carefully pruning redundant ones enhances both generalization and efficiency. Our benchmarking also suggests that certain lightweight architectures are inherently more compatible with spiking conversion, motivating future SNN-oriented redesigns of mobile CNNs. The pruning results further show that simplifying spike pathways can stabilize gradient flow without increasing depth. These insights open opportunities for automated pruning or NAS-driven SNN search, as well as integrating learnable temporal dynamics to further reduce the accuracy gap. Future work will also explore deployment on neuromorphic hardware for practical validation.

\newpage
{
    \small
    \bibliographystyle{ieeenat_fullname}
    \bibliography{main}
}

\end{document}